\documentclass[letterpaper, 10 pt, conference]{ieeeconf}
\usepackage[letterpaper, left=0.75in, right=0.75in, bottom=0.77in, top=0.78in]{geometry}

\IEEEoverridecommandlockouts                              

\usepackage{graphicx}
\usepackage{amsmath}
\usepackage{xcolor}
\usepackage{floatrow}
\usepackage{epstopdf}
\usepackage[colorlinks = true,
            linkcolor = black,
            urlcolor  = black,
            citecolor = black,
            anchorcolor = black]{hyperref}

\usepackage{algorithm,algpseudocode}
\usepackage[font=small]{caption}
\usepackage{accents}
\usepackage{amsfonts}
\usepackage{wrapfig}
\usepackage{booktabs}
\usepackage{balance}
\usepackage{scalerel}
\usepackage{subfigure}
\usepackage{multirow}
\usepackage{url}
\usepackage{svg}
\usepackage{hhline}
\usepackage{hyperref}

\newcommand{\norm}[1]{\left\lVert#1\right\rVert}

\long\def\ignore#1{}
\DeclareMathOperator*{\argmax}{argmax}
\begin{document}
\title{\LARGE \bf
ProgressLabeller: Visual Data Stream Annotation for Training Object-Centric 3D Perception
}

\author{Xiaotong Chen\hspace{0.5cm} 
Huijie Zhang\hspace{0.5cm}  Zeren Yu\hspace{0.5cm}  Stanley Lewis\hspace{0.5cm} Odest Chadwicke Jenkins
\thanks{\authorrefmark{1}X. Chen, H. Zhang, Z. Yu, S. Lewis and O. C. Jenkins are with the Department of Electrical Engineering and Computer Science, and Robotics Institute at the University of Michigan, Ann
Arbor, MI 48109 USA {\tt\footnotesize [cxt|huijiezh|yuzeren|stanlew|ocj] @umich.edu}}%
\thanks{Digital Object Identifier (DOI): see top of this page.}
}


\maketitle
\begin{abstract}

Visual perception tasks often require vast amounts of labelled data, including 3D poses and image space segmentation masks. The process of creating such training data sets can prove difficult or time-intensive to scale up to efficacy for general use. Consider the task of pose estimation for rigid objects. Deep neural network based approaches have shown good performance when trained on large, public datasets. However, adapting these networks for other novel objects, or fine-tuning existing models for different environments, requires significant time investment to generate newly labelled instances. Towards this end, we propose ProgressLabeller as a method for more efficiently generating large amounts of 6D pose training data from color images sequences for custom scenes in a scalable manner. ProgressLabeller is intended to also support transparent or translucent objects, for which the previous methods based on depth dense reconstruction will fail.
We demonstrate the effectiveness of ProgressLabeller by rapidly create a dataset of over 1M samples with which we fine-tune a state-of-the-art pose estimation network in order to markedly improve the downstream robotic grasp success rates. Progresslabeller is open-source at \href{https://github.com/huijieZH/ProgressLabeller}{https://github.com/huijieZH/ProgressLabeller}

\end{abstract}


\section{Introduction}

Visual perception tasks often require vast amounts of labelled data due to their use of deep neural networks.   Such deep neural networks have outperformed traditional methods in object pose estimation \cite{labbe2020cosypose, he2021ffb6d, hodavn2020bop} when trained on public large-scale datasets \cite{xiang2017posecnn, hodan2017tless, hinterstoisser2012model}. 
However, considering the practice of deploying such systems in real-world robotics applications, such as semantic grasping and manipulation, current pose estimation systems can prove the difficulty of adaptation to different objects and settings without retraining with a customized large-scale dataset.

\begin{figure}
    \centering
    \captionsetup{type=figure}
    \includegraphics[width=\columnwidth]{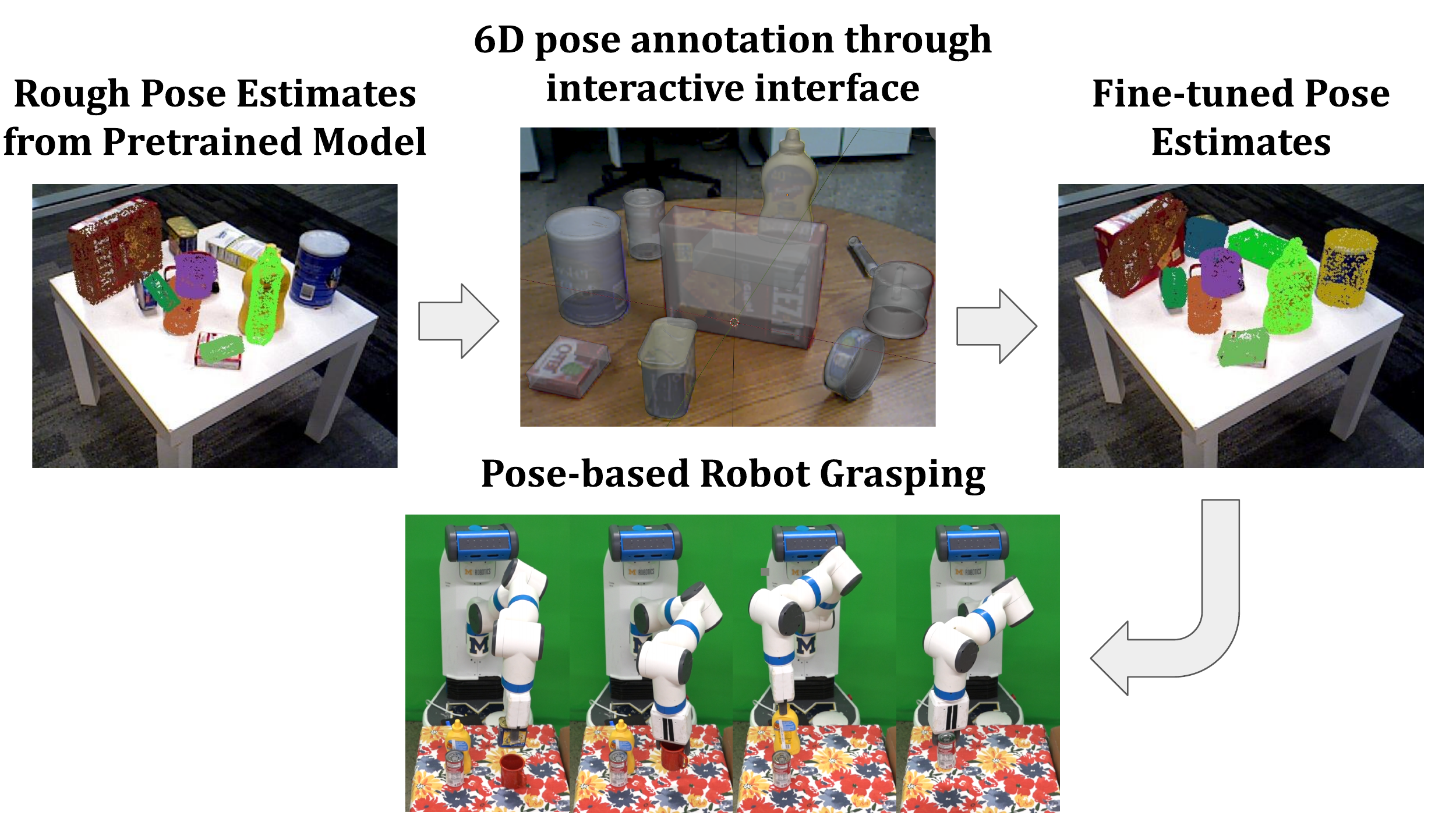}
    \captionof{figure}{The ProgressLabeller offers an interactive GUI for aligning all kinds of objects in the 3D scene to generate large-scale datasets with ground truth pose labels. The left image shows the rough 6D pose estimates from one state-of-the-art RGB-D deep models trained on public YCB dataset, and the right images shows fine-tuned pose estimates from the same model after retraining using generated data from ProgressLabeller. The pose estimates are then used for robotic grasping experiments.}
    \label{fig:teaser}
\end{figure}

In particular, our need for training data is a result of object labels for pose estimation being defined to specific 3D object models (both geometry shape and texture).  Learned models cannot be fine-tuned to transfer to similar object instances without additional training data. Recent work has made advances in category-level or unseen pose estimation \cite{li2020category, park2020latentfusion}. However, the objects included only cover a small set and there is no evidence showing the estimated pose is reliable enough for robotic manipulation. Further, the estimation results of deep neural networks are often vulnerable to environmental changes \cite{chen2019grip}, including different lighting conditions, occlusions and object's special appearance like transparent or reflective surfaces. Synthetic data generation with domain randomization and photo-realistic rendering \cite{to2018ndds, denninger2019blenderproc} could improve generalizability, but it is still challenging to simulate real-world lighting as well as the noise inherent in the sensor modality. We show in the experiment that the network trained using real data is still over-performing synthetic data.

To address the problem of adaptation for deep pose estimation systems and their application to robotic manipulation, we propose \textbf{ProgressLabeller} as a method and implementation for creating large customized datasets more efficiently. 
Inspired by LabelFusion \cite{marion2018label}, ProgressLabeller collects training data of objects {\it in situ} in a mixed-initiative manner, similar in spirit to work by Gouravajhala et al.~\cite{gouravajhala2018eureca}.  It takes visual streams of color images that observe objects in a physical environment as input.  Objects in this stream only need to be labelled once by a human user through visual annotation. ProgressLabeller builds on recent advances in Structure-from-Motion \cite{schonberger2016structure} and visual SLAM \cite{mur2017orb} to produce both a 3D reconstruction and camera pose along the trajectory of the collected visual stream, where the  annotated object labels can be propagated to all frames.   

Compared to depth-based fusion methods, the color feature-based pipeline of ProgressLabeller suffers less noisy or invalid readings than that from depth sensing.  Further, the use of color by ProgressLabeller allows it to include objects that are transparent and reflective \cite{liu2021stereobj} into the pose estimation process,  as long as there exists textures from other objects or background.  From an interface perspective, our implementation of  ProgressLabeller aims to provide a more interactive design geared for users performing labeling tasks.  This interface design enables seamless labelled pose validation in different views by checking and correcting the discrepancy between the masks of re-projected object models and original RGB images (an example is shown in Figure \ref{fig:blender_view}). The intuition is that, if then the labelled pose is close to ground truth, the re-projected object mask should align with the object's true area in RGB images from multiple views.

In this paper, we introduce ProgressLabeller as a semi-automatic approach to object 6D pose labelling on RGB(D) image sequence/video frames that works for transparent objects.  Our aim is to release ProgressLabeller as an open-source tool for more effective dataset generation for object and pose recognition by robots.
We evaluate the labelling accuracy of ProgressLabeller against LabelFusion on data stream samples from 4 public datasets with respect to segmentation mask and object pose accuracy. With the proposed system, we created a dataset of YCB objects \cite{calli2015ycb} (about 1.2M object instance labels) within 2 days of data collection and labelling. The dataset presents more challenges than the public YCB-Video dataset \cite{xiang2017posecnn} with more occlusions and better coverage of camera view directions, and collected using three different RGB-D cameras to evaluate the generalization across sensors. We fine-tuned a state-of-the-art RGB-D deep pose estimator \cite{he2021ffb6d} on our dataset and observed a large improvement on pose estimation accuracy and robotic grasping success rate, compared with the pretrained model on public dataset, as well as on the same amount of data from image-synthesis or LabelFusion annotation.  


\section{Related Work}
With ProgressLabeller, a user can scalably label new datasets with camera world pose, scene object poses and scene object segmentations. This process is enabled by fusing streaming RGB (or RGB-D) inputs into a single scene-wide representation, and then allowing a human user to input relevant 6-DoF information via 3D modelling interfaces (such as those provided by Blender \cite{blender}). This process demonstrates label stability even over long input video streams, and due to its functionality with direct RGB inputs, can label even difficult objects such as transparent cups.  We discuss below methods related to ProgressLabeller.

\subsection{Direct \& Human-in-the-loop labelling}
The creation of 2D segmentation data is analogous to the object detection, keypoint detection, or semantic segmentation tasks (depending on desired output labels). Tools such as LabelMe \cite{russell2008labelme} required users to directly interact with the underlying data to be labelled. This manual process was improved by model-assisted approaches such as Deep Extreme Cut \cite{maninis2018deep} which decreases the amount of user effort necessary to label images.
Shared autonomy and mixed-initiative methods have also been used in this approach, in which the user provides coarse pose or other estimations which are fine-tuned via a model-informed approach \cite{ye2021human}.

 
\subsection{End-to-End Labellers}
Previous tools have been created to enable this style of learning process. LabelFusion \cite{marion2018label} is perhaps the most commonly utilized example. LabelFusion utilizes streaming RGB-D inputs to create a dense reconstruction of the scene, which is then labelled semi-manually by aligning 3D object models to the 3D reconstruction. While this approach is typically robust, it relies on RGB-D input for reconstruction, and experiences difficulties under certain regimes. In particular, transparent objects cause problems for commonly employed depth sensor technologies, and long-running input streams typically result in large amounts of 'drift'. 

Some methods have been introduced to eliminate the need for CAD models in the labelling process. Singh et al. \cite{singh2021rapid} proposed a method which utilizes user labelled keypoints and bounding boxes to generate pose and segmentation labels. This frees the system from dependency on CAD models, but requires user interaction directly with the images. SALT \cite{stumpf2021salt} proposed utilizing GrabCut to generate 3D bounding boxes and image segmentation labels for relevant scenes. This allows removing the dependency on object masks while also allows the labelling of dynamic scenes such as human gait videos. 

Other works sought to improve the labelling procedure itself. EasyLabel \cite{suchi2019easylabel} allows for semi-automatic labelling of scenes via sequentially added objects.  This process is scalable, and generates high quality labels. However, it requires tight physical control over the scene to be labelled, which is not always feasible to obtain. Objectron \cite{ahmadyan2021objectron} utilized modern smartphone's AR capabilities combined with human-labelled 3D bounding boxes to scalably create a large scale dataset. This method however is susceptible to label drift during long-duration input videos. KeyPose \cite{liu2020keypose} specifically sought to generate labelled datasets for transparent objects. This method utilized stereoscopic images taken from a robot armature in order to avoid the problems of typical depth cameras have with transparent objects.

\section{ProgressLabeller}

In ProgressLabeller, we provide an interactive GUI for users to label object poses for a large amount of data in several steps. We incorporate camera pose estimation systems that reconstructed a 3D scene for aligning objects with, and calculate pose transforms that can be use to propagate the labelled object pose to all image frames. We provide interface to switch views among every RGB images to enable verification of labelled object pose by checking the alignment of re-projected mask with RGB images in multiple views.
In this work, we assume the 3D mesh models of objects are provided and the objects are static in the scene during data collection (in special cases like T-LESS dataset~\cite{hodan2017tless}, the objects are put onto a turntable and moving altogether while the cameras are static, the algorithm can still work because no environment background is captured by the cameras).
\subsection{Camera pose estimation}
We incorporated ORB-SLAM3 \cite{campos2021orb}, KinectFusion \cite{newcombe2011kinectfusion} and COLMAP \cite{schonberger2016structure, schonberger2016pixelwise} to do RGB, depth or RGB-D camera pose estimation as well as reconstruction. ORB-SLAM3 is selected as the default method for its balanced speed and accuracy. COLMAP was tested and proved to be more accurate but the time cost of reconstruction over more than 1000 images is unaffordable.

\subsection{Object alignment by multi-view silhouette matching}
\begin{figure}[htbp]
     \centering
     \includegraphics[width=0.95\columnwidth]{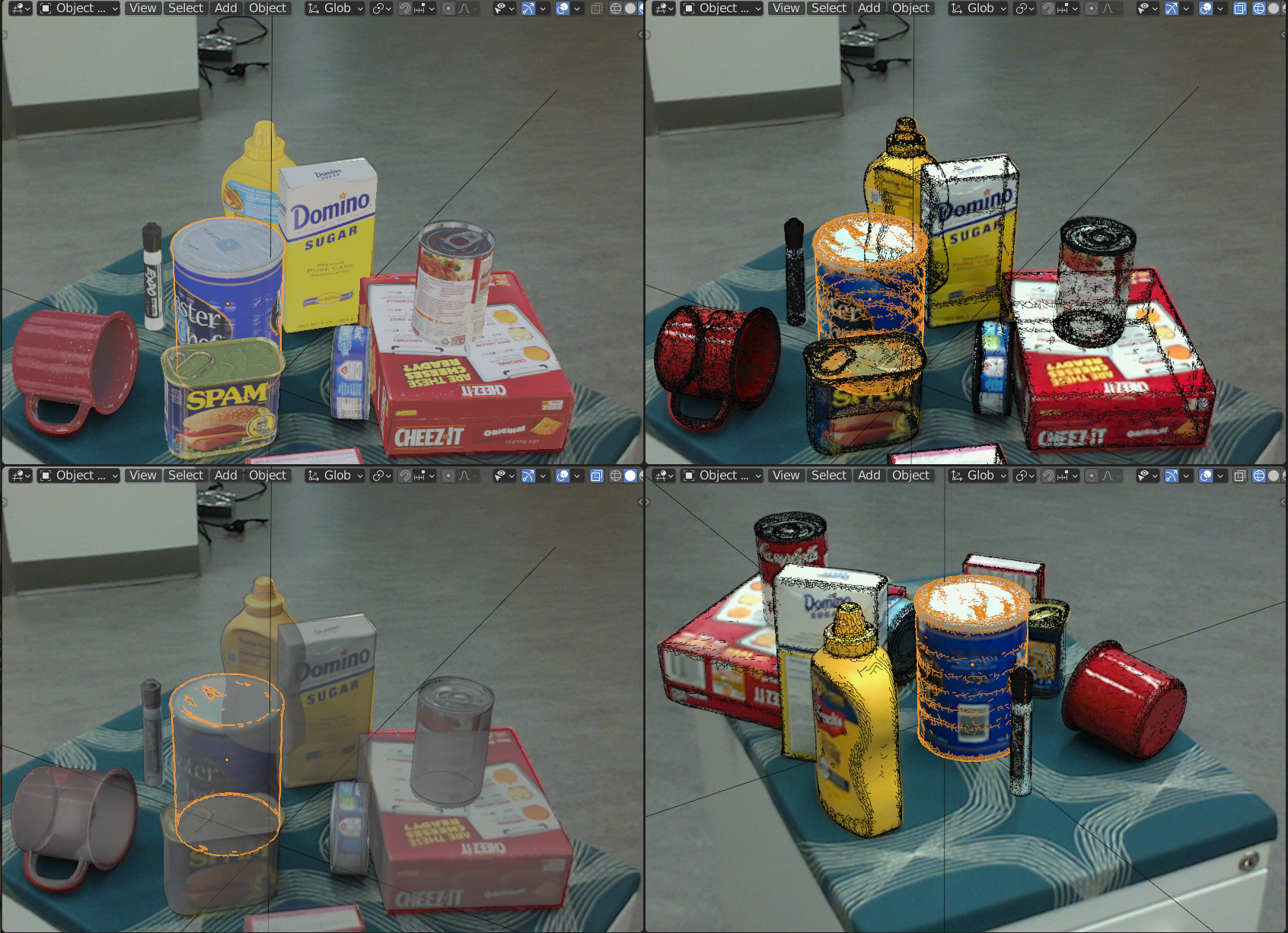}
     \caption{Capture of ProgressLabeller user interface for multi-view re-projection checking. The top-left view shows the aligned object models with rendered texture at labelled poses. The bottom-left view and top-right view show the silhouettes and boundaries respectively at the same camera view, and the bottom-right view shows the boundaries from another view for validation.}
     \label{fig:blender_view}
 \end{figure}

Different from methods that align object 3D models with reconstructed point clouds, our system created a multi-view graphical user interface that overlays the object model's projection onto the original RGB images, so that the object pose errors could be easily detected from areas with misalignment of object texture, silhouette and boundary, as shown in Figure~\ref{fig:blender_view}. Compared to depth reconstructed based methods, the pose label accuracy is improved based on higher accuracy of RGB than depth sensing. Besides, the system can also be used to label scenes with unreliable depth from transparent objects and backgrounds (see Section~\ref{sec:other}).

\subsection{Semi-automatic labelling pipeline}

We build ProgressLabeller as a plugin on Blender \cite{blender}, which provides a good multiple view interface of overlaid RGB image, 3D reconstructed scene and objects for labelling and verifying poses. The overall procedure of labelling object poses includes several steps within the Blender graphical interface: 
\begin{enumerate}
    \item Import RGB(D) images and object 3D models into the 3D interactive workspace. Set parameters including camera intrinsics, camera pose estimation parameters and display settings, etc.
    \item Do camera pose estimation, then the estimated camera poses and a reconstructed 3D point cloud will be added to the workspace. Depth images are used to solve the scale of reconstruction.
    \item (optional) When depth input is available, another point cloud fusing depth input based on estimated camera poses can be generated to provide a denser view of the entire scene and help find objects' rough positions.
    \item (optional) Do plane alignment for a better viewpoint and ease of correcting object pose
    \item Align the object so that its re-projection matches the ground truth area in multiple views of RGB images
    \item Export labelled object poses and render segmentation masks, bounding box labels, etc.
\end{enumerate}

\noindent
In step 2, when there is only RGB image available, the scene scale (3D point cloud and camera's trajectory position) is unknown, to solve this scale problem, we take advantage of the known object sizes by dragging them to align with the 3D reconstruction and verify that in multiple different views.
In step 3, we implemented a depth fusion module based on the estimated camera poses from RGB reconstruction, based on the Signed Distance Functions as in KinectFusion \cite{newcombe2011kinectfusion}. The fused point cloud could give a rough reference of object locations.
In step 4, Iterative Closest Point (ICP) is used for aligning RGB feature or depth point cloud to X-Y plane in Blender.
In step 5, the object pose is labelled with visually ensured accuracy, which is the essential design that enables labelling of objects with only RGB images based on multi-view geometry. By re-projecting object's 3D model and check whether it aligns perfectly with the object's true textures or silhouettes in the RGB images from multiple views, users can fine-tune the object's pose and verify the error seamlessly until the object matches the images. 

The entire labelling pipeline typically takes around 30 minutes for one data stream about 5K images. Data import and export takes about 20-30\% of time, with a rendering speed around 4-20 images per second depending on number of objects in the scene. The rendering could be done in parallel on a GPU if a vast amount of data is required. ORB-SLAM3 camera pose estimation takes about 10-20\% of time. Manual labelling and verification take the rest of time. 
\subsection{Annotation accuracy estimation from simulation}

We verify the accuracy of annotations throughout this multi-view silhouette matching process by simulating an iterative object pose update process. In each iteration, given a certain camera frame, 
we assume the object will be translated in a plane parallel with its x-y plane, or rotated about z-axis (for better control), towards a pose that maximizes the Intersection-over-Union (IoU) between the rendered silhouette at current pose and ground truth.

\subsubsection{Problem Definition} Given set of $N$ images $I^{\{i\}}$ with their corresponding camera pose $T^{\{i\}}$ in the world frame, $i \in \{1, 2, 3, \ldots, N\}$. Our goal is to find ground truth object pose $T^{\text{obj}\{j\}}_{gt}$ in the world frame for all the objects $j \in \{1, 2, 3, \ldots, M\}$ in the scene. We define the projection operator as $S^{\{i, j\}} = \text{Proj}(T^{\{i\}}, T^{\text{obj}\{j\}})$, which render object $j$ given its CAD model, camera pose $T^{\{i\}}$ and object pose $T^{\text{obj}\{j\}}$ into an object texture/silouette $S^{\{i, j\}}$. Also defined the IoU operator as $\text{IoU}_{\text{obj}\{j\}}(I^{\{i\}}, \text{Proj}(T^{\{i\}}, T^{\text{obj}\{j\}}))$ to calculate the IoU for pixels in object $j$ between real image $I^{\{i\}}$ and synthetic texture/silouette $S^{\{i, j\}}$. 

\begin{figure}[htbp]
     \centering
     \includegraphics[width=0.8\columnwidth]{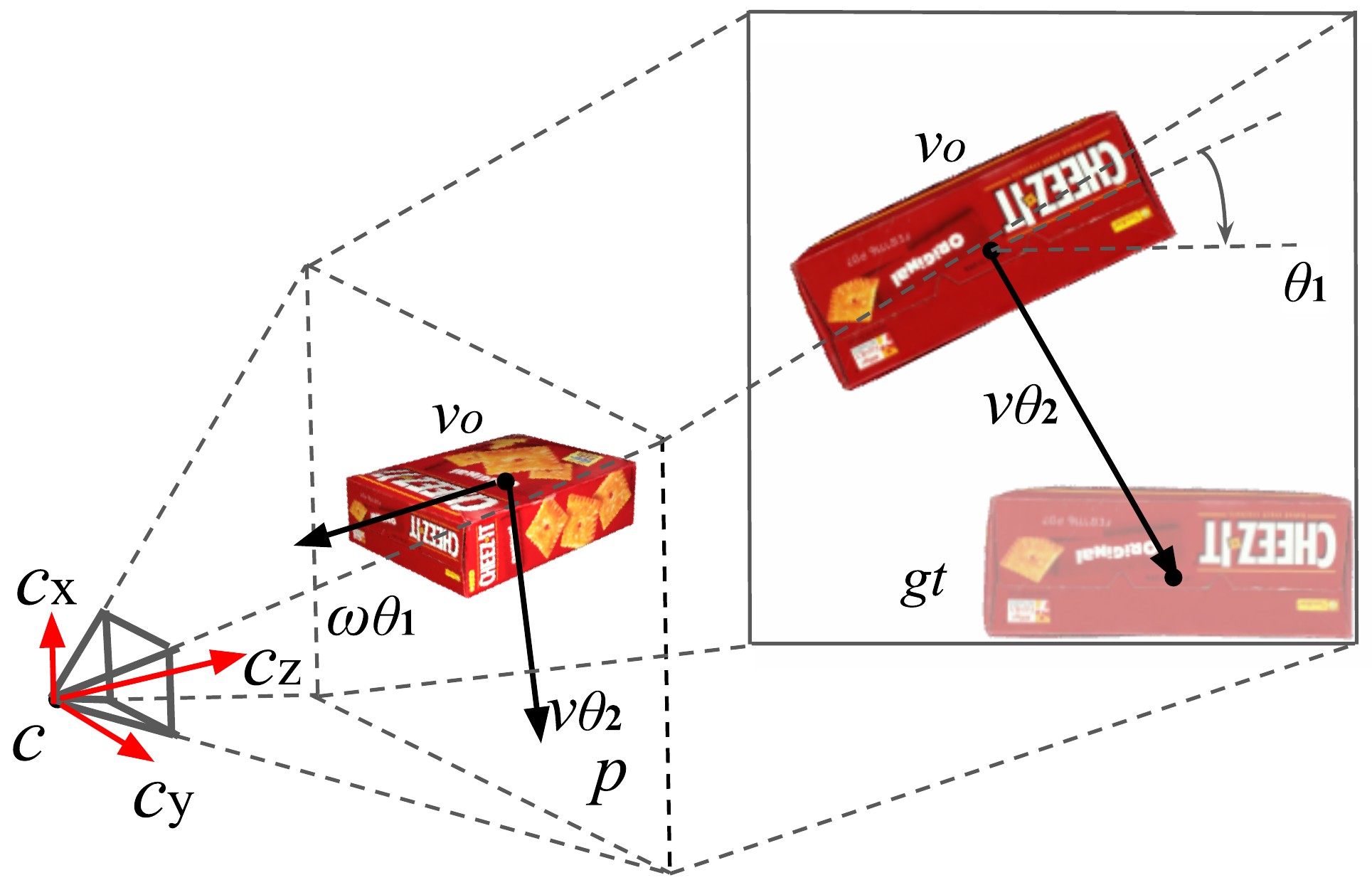}
     \caption{Diagram for an object shown under a camera. $c$ denote the location of camera and $c_x, c_y, c_z$ are its x, y, z axis. $p$ is a plane parallel to camera plane and passing through object's center $v_o$. $\omega$ the rotation direction parallel to $c_x$ and passing through $v_o$. $\theta_1$ is the magnitude of rotation radius. $v$ is the translation direction within the plane $p$ and $\theta_2$ is the translation magnitude. On the right hand side is the projection image, the object in the transparent color is the object with ground truth pose.}
     \label{fig:annotation_limitation}
 \end{figure}

The multi-view texture/silhouette matching iterative update is proceeded with a goal to maximize the IoU. Given the pose for object $j$ in the $k$th iteration $T^{\text{obj}\{j\}}_{(k)}$, in ($k + 1$)th iteration:

\begin{equation}
    T^{\text{obj}\{j\}}_{(k+1)} = \argmax_{f[T^{\text{obj}\{j\}}_{(k)}]}\text{IoU}_{\text{obj}\{j\}}(I^{\{i\}}, \text{Proj}(T^{\{i\}}, f[T^{\text{obj}\{j\}}_{(k)}] )) 
    \label{eq:iterate}
\end{equation}
where $f[T^{\text{obj}\{j\}}_{(k)}]$ describes all possible translation start from the  $T^{\text{obj}\{j\}}_{(k)}$ 
that is within the plane $p$ or the rotation along the axis $\omega$ as shown in Figure \ref{fig:annotation_limitation}. So:

\begin{equation}
    f[T^{\text{obj}\{j\}}_{(k)}] = \exp^{\widehat{\xi}_1 \theta_1}  \exp^{\widehat{\xi}_2 \theta_2} T^{\text{obj}\{j\}}_{(k)}
    \label{eq:range}
\end{equation}
where ${\xi}_1 = \begin{bmatrix}-\omega \times v_o \\ \omega \end{bmatrix}$, ${\xi}_2 = \begin{bmatrix} v \\ 0 \end{bmatrix}$ are the twist coordinates for twist $\widehat{\xi}_1, \widehat{\xi}_2$.




\subsubsection{Simulation results}

We generate a CAD model sets with 44 different CAD models. For each run, we generate $T^{\text{obj}}_{gt}$ with a random rotation matrix and location at the origin. 40 cameras are created with their z axis pointing towards the origin and a random location at a sphere around the object. The initial pose $T^{\text{obj}}_{0}$ is generated by adding a random position noise from Gaussian distribution with variance of 10cm to origin and with a random 3D orientation. During each iteration, $v, \theta_1, \theta_2$ in Equation \ref{eq:range} are discretized for simulation. The result shows that it takes around 10.36 iterations for the algorithm to converge within 1mm location error and dot product larger than 0.99 between ground truth and converged rotation axes.

\section{Experiments}
\begin{figure*}[htbp]
     \centering
     \includegraphics[width=\textwidth]{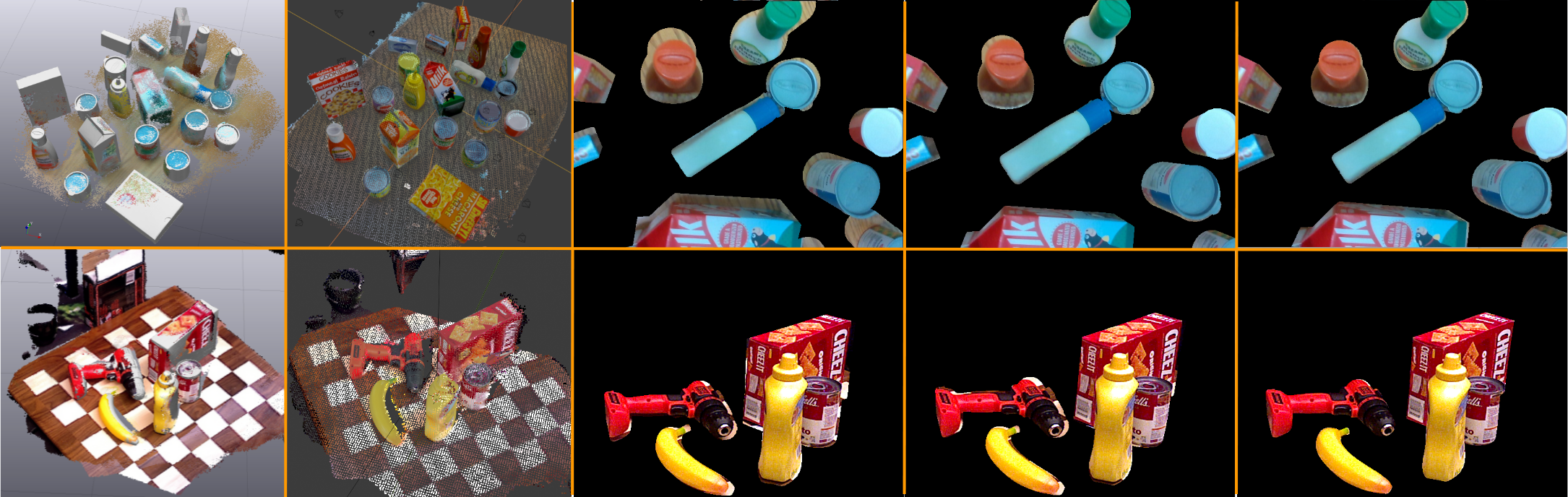}
     \caption{From left to right, we show the 3D labeller view of LabelFusion, ProgressLabeller and cropped RGB image with masks from LabelFusion, ground truth and ProgressLabeller. The top and bottom row show a sample from HOPE and YCB dataset respectively. We find even ground truth pose labels have easily observable errors, shown as the spacing between black masks and object boundaries.}
     \label{fig:compare_label}
 \end{figure*}

In this section, we report results on several evaluation experiments on the data generated using ProgressLabeller. The output label quality is firstly evaluated on public datasets against ground truth. Then we introduce a large scale dataset with object pose annotations, and report the evaluation results of fine-tuning a state-of-the-art deep pose estimation on accuracy. The improved accuracy is further evaluated in robotic grasping experiment. Finally, we show two potential applications: labelling a complex scene with transparent objects in front of reflective backgrounds, and training a neural rendering model on a single object from collected data and do image synthesis.

 \begin{table*}[htbp]
\small
   \centering
   \begin{tabular}{@{}p{2cm}p{2.2cm}p{1cm}p{1cm}p{1.2cm}p{1.8cm}p{1.8cm}p{1.4cm}@{}}
     \toprule
     Dataset & Method & Mask IoU $\uparrow$ & Pos (cm) $\downarrow$ & Rot (quat) $\downarrow$ & ADD \hphantom{1cm} (AUC-0.1d) $\uparrow$ & ADD-S (AUC-0.1d) $\uparrow$ & Feature (Pixel) $\downarrow$\\
     \midrule
     HOPE & LabelFusion$^*$    & 0.6729  & 1.5825 & 0.2561 & 0.7095 & 0.7101 & 4.1248\\
      & ProgressLabeller     & 0.8600  & 1.0948 & 0.0801 & 0.9566 & 0.9566 & 1.9479\\
      & Ground Truth        & - & - & - & - & - & 2.2752\\
     \midrule
     YCB-Video & LabelFusion  & 0.9089 & 0.5165 & 0.0741 & 0.9977 & 0.9989 & 4.7365\\
      & ProgressLabeller      & 0.8849 & 1.7537 & 0.0636 & 0.7642 & 0.7662 & 3.08\\
      & Ground Truth         & - & - & - & - & - & 3.3868\\
     \midrule
     NOCS$^*$ & LabelFusion      & 0.7684 & 2.2825 & 2.9092 & 0.6436 & 0.6438 & -\\
      & ProgressLabeller      & 0.8785 & 1.4949 & 2.6153 & 0.8674 & 0.8679 & -\\
     \midrule
     T-LESS$^*$ & LabelFusion*    & - & - & - & - & - & -\\
      & ProgressLabeller      & 0.8810 & 0.6779 & 0.0856 & 0.6626 & 0.6642 & -\\
     \bottomrule
   \end{tabular}
   \caption{Results on feature matching distance between rendered object RGB image and original image. 'Pos (cm)', 'Rot (quat)' refers to positional error in centimeters and rotational error calculated as norm of difference in quaternions as in \cite{huynh2009metrics}. LabelFusion cannot work on T-LESS data streams at all, and it cannot reconstruct one of HOPE samples scene entirely, so we trim the sequence to 1/4 length. Object models in T-LESS and NOCS datasets do not have enough features to measure the feasure-based pixel distances.}
   \label{tab:example}
 \end{table*}

\subsection{Evaluation on label generation}
We evaluate the label quality with respect to pose accuracy against ground truth and time cost on ProgressLabeller versus LabelFusion \cite{marion2018label}. Specifically, we use both tools to label on 64 object instances from 8 sample sequences among the 4 public pose estimation datasets including YCB-Video \cite{xiang2017posecnn}, T-LESS \cite{hodan2017tless}, NOCS \cite{li2020category}, and HOPE \cite{hope_github}.

\noindent
\textbf{Evaluation metrics.} We use mask Intersection-over-Union (IoU) between rendered mask of object at labelled pose and at ground truth to evaluate the segmentation, and select 4 distance metrics to compare a labelled pose with ground truth pose, including 3D positional error $E_p$, 3D rotational error $E_r$, average pairwise distance (ADD) $E_{ADD}$ \cite{hinterstoisser2012model} and average pairwise distance in symmetric case (ADD-S) $E_{ADDS}$ \cite{xiang2017posecnn}, defined as follows:

\begin{align}
    &E_p = \norm{t-t^*}, E_r = \min \{\norm{q - q^*}, \norm{q + q^*} \}\\
    &E_{ADD} = \dfrac{1}{|\mathit{M}|} \sum_{x \in \mathit{M}} \norm{(Rx + t) - (R^* x + t^*)}\\
    &E_{ADDS} = \dfrac{1}{|\mathit{M}|} \sum_{x_1 \in \mathit{M}} \min_{x_2 \in \mathit{M}}\norm{(Rx_1 + t) - (R^* x_2 + t^*)}
\end{align}
where $\mathit{M} = \{x_i \in \mathbb{R}^3\}$ is the object 3D point set, and $R, q, t, R^*, q^*, t^*$ refers to the 3D rotation matrix, rotation represented in quaternion, 3D position vector of labelled pose and ground truth respectively. Specifically for ADD and ADD-S, we report the value of area under the accuracy-threshold curve obtained by varying the distance threshold from 0 to 0.1 diameter of objects (AUC-0.1d) following metrics among pose estimation papers.

Besides, from the observation that even ground truth labels are still not perfectly correct (re-projected object masks align with true area in RGB images from multi-view), we propose an RGB feature based evaluation on object pose without ground truth. 
Specifically, we render the object at labelled pose to get an image with only an object in front of black background, and extract SIFT features from both the rendered image and original RGB image, then calculate the average of feature matching distance in pixels. 
In practice, we assume the labelled pose is close to ground truth, and remove incorrect matching with a distance threshold of 10 pixels. Figure \ref{fig:compare_label} shows qualitative examples of labelled poses, where we observe more accurate re-projection results from ProgressLabeller than LabelFusion output as well as provided ground truth on masked RGB images. We believe the main error cause in LabelFusion is the accumulated depth sensing error during reconstruction. Another drawback of LabelFusion UI design is that, there is no measurement or display of object poses in the 3D scene, so the users need to switch back-and-forth between the 3D scene and the final generated images that has image masks rendered at labelled poses. Also its 3D model as well as rendering result doesn't display textures, so it's hard to deal with symmetric-shape objects in HOPE and NOCS. 

The quantitative results are shown in Table~\ref{tab:example}. From the comparison, we see  ProgressLabeller is more accurate and robust in most streams. LabelFusion has higher accuracy on YCB-Video dataset, which has a large feature-based pixel distance. For example, we found the pose annotation of object 006\_mustard\_bottle was flipped 180$^\circ$ in one scene. Besides, the two approaches took similar annotation time of 10-30 minutes per scene.

\subsection{Multi-camera dataset creation}
\begin{figure}[htbp]
     \centering
     \includegraphics[width=\columnwidth]{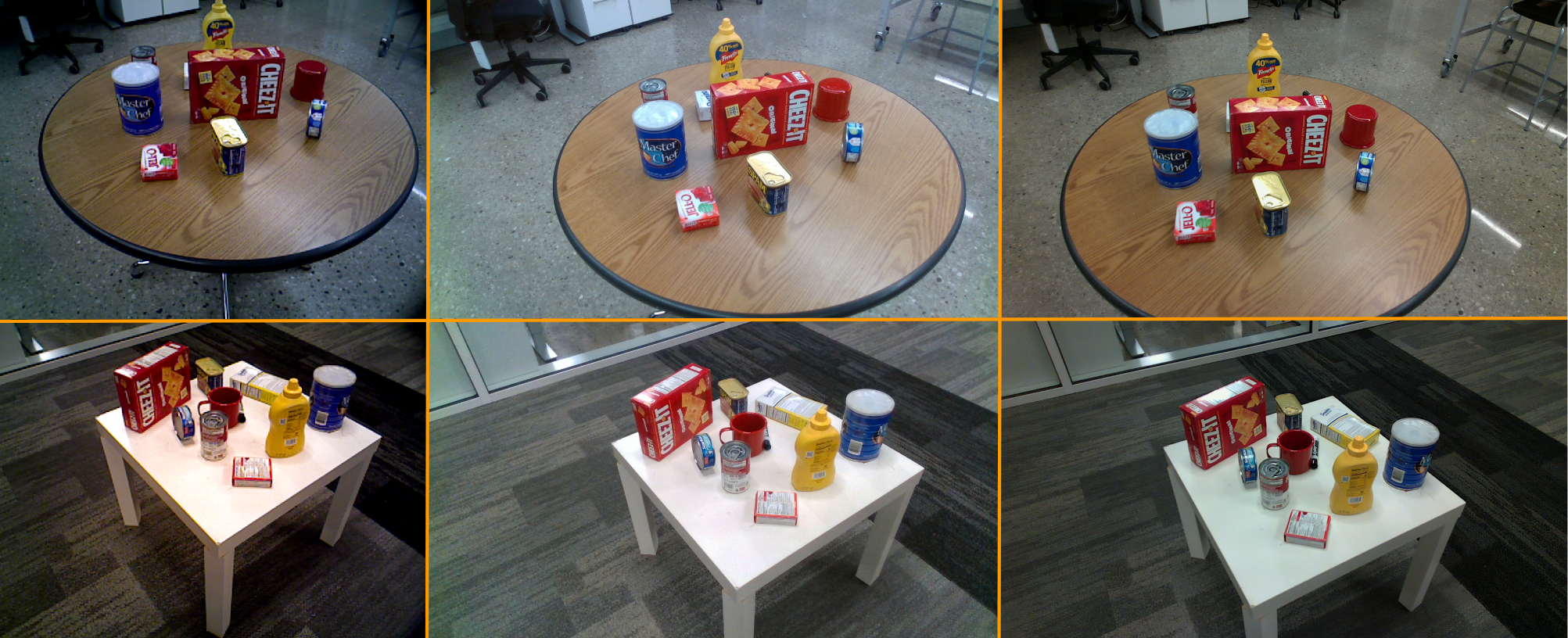}
     \caption{The top and bottom row shows RGB images in training and testing set respectively. From left to right, the images are taken from Primesense, RealSense D435i and L515.}
     \label{fig:dataset}
 \end{figure}

From analysis of absolute accuracy result above and to evaluate robustness on existing deep pose estimation networks, we aim to create a more accurate, multi-camera, cluttered dataset on YCB objects \cite{calli2015ycb}. We mounted 3 RGB-D cameras with different sensing technologies, ASUS Xtion Pro Primesense Carmine 1.09 (structured light), Intel RealSense D435i (stereo) and RealSense L515 (LiDAR), on a Fetch robot and collected data streams at $\sim$15Hz when Fetch is slowly driving around the tabletop object scene with a constant speed. Each scene contains almost a bit more than full round, with 1K$\sim$3K paired RGB-D images from each camera. In this way, the collected data is free of motion blur and covers full round view of objects and occlusions. The dataset includes 16 scenes with 10 objects placed in both isolation and dense clutter, and another 4 scenes each with a single object for unit test. Figure \ref{fig:dataset} gives an example. It took 2 days collected and labelled the dataset, with about 120K images and 1.2M labelled object instances.

 \begin{table*}
   \centering
\small
   \begin{tabular}{@{}lcccccc@{}}
     \toprule
     Train set/Test set & \multicolumn{2}{c}{Primesense} & \multicolumn{2}{c}{RealSense L515} & \multicolumn{2}{c}{RealSense D435i} \\
     \midrule
     Metric: AUC (0-10cm)& ADD & ADD-S & ADD & ADD-S & ADD & ADD-S \\
     \midrule
     Pretrained on Asus Xtion Pro Live & 21.63 & 41.16 & 47.43& 72.90 & 44.67&	68.00\\
     Fine-tuned on Primesense &  \textbf{60.02} & 	\textbf{79.21}& 55.32 & 77.85 & 49.75 & 73.35\\
     Fine-tuned on RealSense L515 & 33.45 &	58.16 & \textbf{68.84} &\textbf{83.69}  & 57.67 & 76.26\\
     Fine-tuned on RealSense D435i & 26.39 & 50.38 & 63.82& 82.52 & \textbf{64.64}	&\textbf{82.36}\\
     \bottomrule
   \end{tabular}
   \caption{The pose estimation accuracy of FFB6D cross-validated on datasets collected using three cameras.}
   \label{tab:add}
 \end{table*}

\subsection{6D object pose estimation fine-tuning}
Using the collected dataset, we fine-tuned one of the state-of-the-art deep pose estimation neural networks taking RGB-Depth images as input, FFB6D \cite{he2021ffb6d} over its pretrained model on YCB-Video dataset. In particular, we created 3 training sets by downsampling to 1/10 the training set collected by 3 cameras respectively and got 3 fine-tuned models by training on a RTX 3080 with batch size 6 for 20 epochs and default settings in other parameters. Then, we cross-validated the 3 fine-tuned models, along with pretrained model, on 3 test sets taken from 3 cameras.

\subsubsection{Evaluation on pose accuracy across different cameras}
We report the AUC for ADD and ADD-S metrics between 0 and 10cm, to match the original result in \cite{he2021ffb6d}. The result is shown in Table \ref{tab:add}, where we find the fine-tuned dataset recognizes the sensor modality and noise difference between 3 cameras as the test set ADD and ADD-S are mostly the highest on the same training set. 

\begin{figure}[htbp]
    \centering
    \includegraphics[width=0.95\columnwidth]{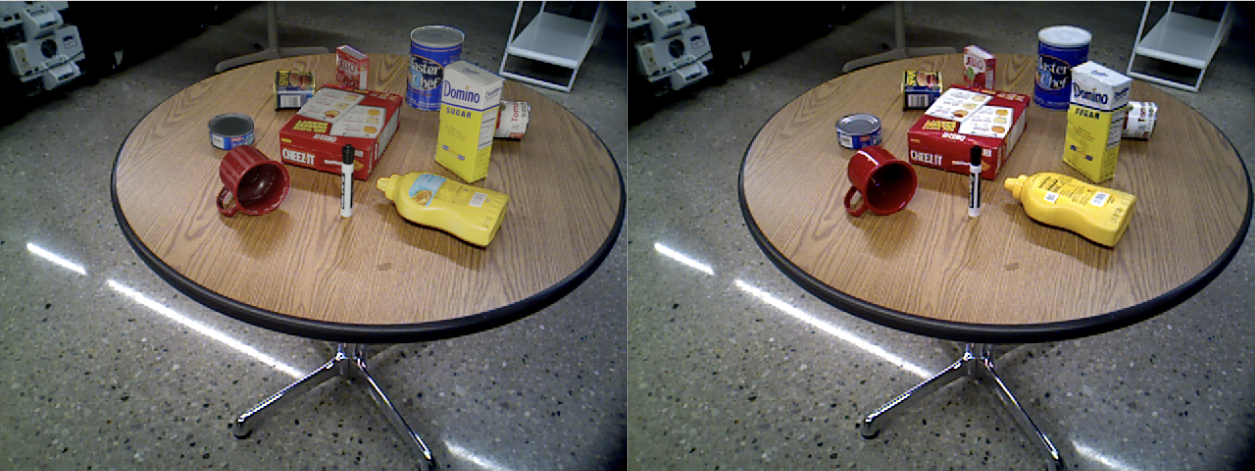}
    \caption{A synthesized RGB image using Blenderproc is shown on the left. The original image is shown on the right. The YCB objects are rendered and overlaid to original image according to labelled poses. Depth synthesized images are generated in the same way.}
    \label{fig:bp}
\end{figure}

\subsubsection{Evaluation against LabelFusion and synthetic data on pose accuracy and robotic grasping} We used 3 datasets to train FFB6D with the same setting as above. The first is the collected Primesense subset labelled by ProgressLabeller, the second is the same data labelled by LabelFusion, and the third is synthesized images by rendering YCB objects over both raw RGB and depth images according to the labelled poses using Blenderproc~\cite{denninger2019blenderproc}, which mostly maintain the same object pose distribution and sensor noise as real data, as shown in Figure~\ref{fig:bp}. 

\begin{table*}
  \centering
\small
  \begin{tabular}{@{}lccccccccc@{}}
    \toprule
    \multicolumn{2}{c}{Network model}  & \multicolumn{2}{c}{Pretrained} & \multicolumn{2}{c}{Blenderproc}& \multicolumn{2}{c}{LabelFusion} & \multicolumn{2}{c}{ProgressLabeller} \\
    \midrule
    Object name & Grasp tolerance (cm) & ADD & Grasp   & ADD & Grasp  & ADD & Grasp   & ADD & Grasp \\
    \midrule
    002\_master\_chef\_can      & 0.36 & 25.16 & 0/5 & 48.67 & 3/5  & 44.78 & 4/5 & 55.42 & 5/5\\
    \midrule
    003\_cracker\_box           & 3.90 & 19.57 & 1/5 & 50.76 & 1/5  & 60.8 & 4/5 & 71.01 & 5/5\\
    \midrule
    005\_tomato\_soup\_can      & 3.65 & 27.74 & 1/5 & 51.12 & 2/5  & 65.48 & 4/5 & 66.79 & 5/5\\
    \midrule
    006\_mustard\_bottle        & 6.33 & 31.02 & 1/5 & 58.87 & 5/5  & 70.77 & 5/5 & 73.47 & 5/5\\
                                & 2.50 & 31.02 & 0/5 & 58.87 & 5/5  & 70.77 & 4/5 & 73.47 & 5/5\\
    \midrule
    007\_tuna\_fish\_can        & 7.09 & 15.14 & 2/5 & 26.41 & 3/5  & 30.02 & 5/5 & 39.87 & 5/5\\
                                & 2.04 & 15.14 & 2/5 & 26.41 & 4/5  & 30.02 & 4/5 & 39.87 & 5/5\\
    \midrule
    009\_gelatin\_box           & 7.60 & 13.25 & 2/5 & 32.79 & 4/5  & 51.45 & 5/5 & 55.24 & 5/5\\
                                & 3.14 & 13.25 & 2/5 & 32.79 & 3/5  & 51.45 & 5/5 & 55.24 & 5/5\\
                                & 1.62 & 13.25 & 2/5 & 32.79 & 4/5  & 51.45 & 5/5 & 55.24 & 4/5\\
    \midrule
    010\_potted\_meat\_can      & 4.75 & 19.68 & 5/5 & 39.56 & 3/5  & 47.48 & 4/5 & 50.64 & 5/5\\
                                & 2.08 & 19.68 & 1/5 & 39.56 & 4/5  & 47.48 & 4/5 & 50.64 & 4/5\\
                                & 0.78 & 19.68 & 2/5 & 39.56 & 0/5  & 47.48 & 5/5 & 50.64 & 5/5\\
    \midrule
    025\_mug                    & 1.24 & 19.10 & 1/5 & 60.70 & 3/5  & 49.31 & 1/5 & 66.53 & 4/5\\
                                
    \midrule
    040\_large\_marker          & 8.67 & 20.10 & 0/5 & 44.75 & 2/5  & 55.89 & 3/5 & 53.73 & 3/5\\
    \midrule
    overall                     & -    & -     & 22/75 & -   & 46/75 & - & 62/75 & - & 70/75\\
    \bottomrule
  \end{tabular}
  \caption{Grasp record comparison on part of YCB objects using grasp poses generated based on pose estimates from pretrained and fine-tuned FFB6D on the collected dataset, with Blenderproc data synthesis, LabelFusion annotations and ProgressLabeller annotations.}
  \label{tab:grasp}
\end{table*}

\begin{figure}[htbp]
     \centering
     \includegraphics[width=\columnwidth]{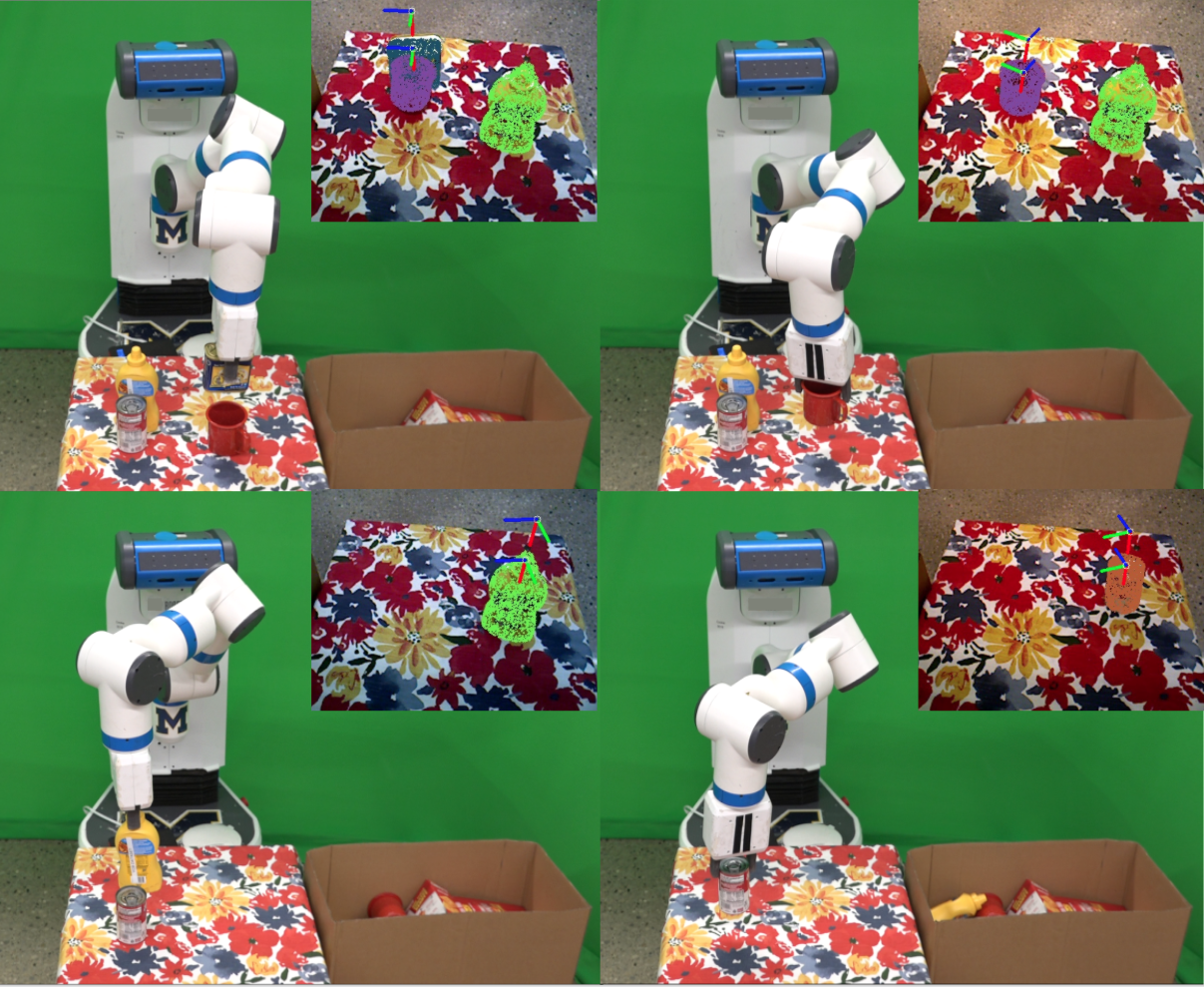}
     \caption{The Fetch robot is grasping objects on tabletop based on estimated poses. The pose estimates are shown as projected point clouds and coordinate frames in top-right smaller images. The higher and lower frames correspond to pre-grasp and grasp poses.}
     \label{fig:grasping}
 \end{figure}

We tested pose estimation accuracy from models trained on the above 3 datasets, with respect to ADD-AUC and pose-based grasping success rate. For pose-based grasping, we manually defined grasp poses along the symmetrical axes on the object 3D models, and grouped them by the grasping tolerance, defined by the distance between two gripper fingers (10.39 cm for Fetch) minus the object's diameter along the grasping direction. Obviously, a smaller grasping tolerance requires more accurate pose estimates for a successful grasp. For example, 002\_master\_chef\_can is challenging to grasp as its tolerance is only 0.36 cm.
We repeated grasping on every tolerance for 5 times based on the pose estimates, and the success/failure statistics is shown in Table \ref{tab:grasp}. Overall, the fine-tuning over ProgressLabeller data improves the grasping success rate the most. Among experiments, we found in some cases the grasp is still successful given large rotational error, where the object was aligned to another pose during grasping, such as 002\_master\_chef\_can and 006\_mustard\_bottle when grasping from the side. Also, sometimes the objects were grasped along another direction, such as 009\_gelatin\_box and 010\_potted\_meat\_can, in this case, their actual grasp tolerance might change. We expect evaluation on object placement accuracy to reveal these errors. Otherwise, the grasping test results generally matches the pose accuracy presented in ADD. Figure~\ref{fig:grasping} shows an example of grasping experiment.

\subsection{Other applications}

\begin{figure*}
    \centering
    \includegraphics[width=0.95\textwidth]{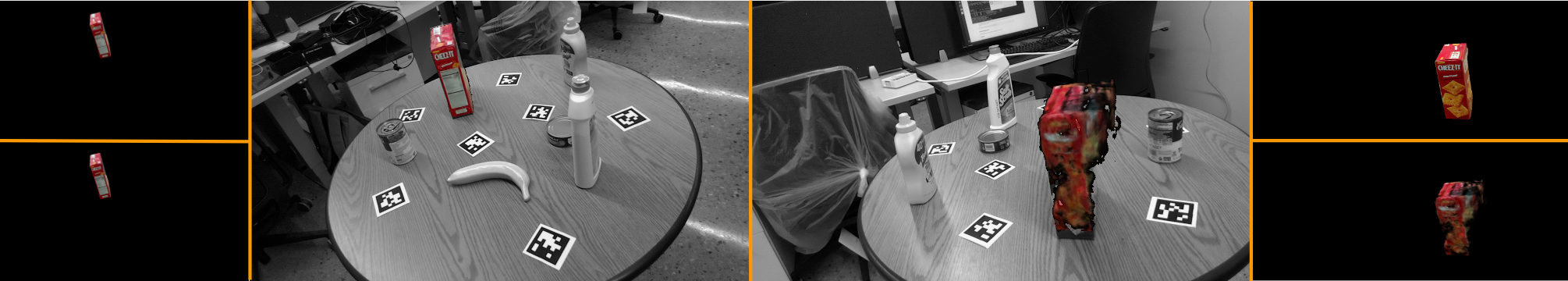}
    \caption{Comparison of best (left) and worst (right) validation pose renderings from the NSVF model. The smaller images with black background on top corners are the ground truth masks, and those on bottom corners are rendered output from NSVF, and the middle larger images are the renderings imposed onto the original captured image in greyscale. The best image (left) had an average pixelwise L2 error of $0.0067$, and the worst (right) had an average pixelwise L2 error of $0.1867$.}
    \label{fig:nsvf_fig}
\end{figure*}

\label{sec:other}
\subsubsection{Transparent Dataset Labelling}
\begin{figure}[htbp]
    \centering
    \includegraphics[width=0.8\columnwidth]{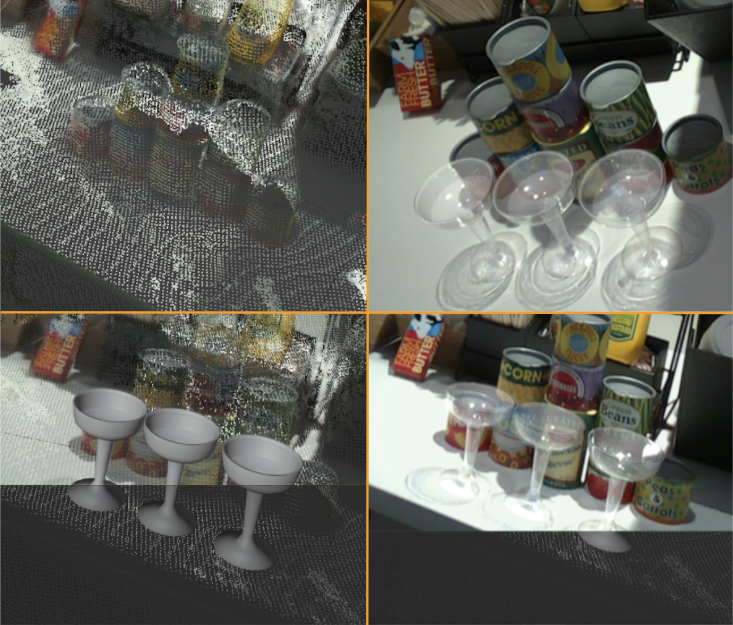}
    \caption{The labelling on transparent champagne cups. When there is no reliable depth (top-left), ProgressLabeller enables checking the pose validity by re-projection object models (bottom-left) to RGB images from different views (top-right and bottom-right).}
    \label{fig:demo_trans}
\end{figure}

We collected RGB-D videos with transparent cups and labelled their poses using ProgressLabeller. Figure \ref{fig:demo_trans} shows in detail that even there is no 3D points around the transparent area when we fused raw depth according to the estimated camera poses, the tool enables accurate pose labelling by matching the object's mask with RGB images. We believe a large-scale 3D dataset of transparent objects can be efficiently created using ProgressLabeller.

\subsubsection{Training neural radiance fields on objects}
As a brief additional experiment, we provide prima facie data to show the effectiveness of our labeller in generating datasets for training a Neural Sparse Voxel Fields (NSVF) model \cite{liu2020neural}. This model permits rendering objects at novel poses given multiple input views with corresponding camera poses. We use ProgressLabeller to label a scene containing a 003\_cracker\_box, with 3509 images in total. From this dataset, we uniformly random sample 100 images with object poses for train, test and validation set respectively. We remove the background from the training images using the segmentation from labeller, and an NSVF model was trained to 75k total iterations on a dual GPU machine consisting of an Nvidia RTX 3060 TI and RTX 3070. After training, renderings were made at each camera pose given in the train, test, and validation sets. These renderings were evaluated via a pixel averaged L2 error against the ground truth segmented images produced by ProgressLabeller where each pixel channel was normalized to the range $[0,1]$. Those results presented in Table \ref{tab:nsvf_metrics}. Visual comparisons between the best and worst validation set renderings along with their associated errors are presented in Figure \ref{fig:nsvf_fig}.

\begin{table}[htbp]
  \centering
  \small
  \begin{tabular}{@{}lccc@{}}
    \toprule
    Dataset & Train & Validation & Test\\
    \midrule
    Per-pixel L2 error & 0.0108 & 0.0460  & 0.0409\\
    \bottomrule
  \end{tabular}
  \caption{Average pixelwise errors for the train, validation, and test sets from the NSVF model trained on ProgressLabeller outputs.}
  \label{tab:nsvf_metrics}
\end{table}

\section{Conclusion}
In this work, we have presented the design and application ProgressLabeller for object pose annotation. Through comparison to LabelFusion, we show that its higher accuracy with similar labelling time, efficiency of creating a large-scale customized dataset, and potential in fine-tuning pose estimation deep neural networks for robotic grasping. In future, we can improve ProgressLabeller by adding online object model generation and support for dynamic scenes.

\noindent
\textbf{Acknowledgement}. We thank greatly the support from Dr. Peter Gaskell and Liz Olson at the University of Michigan, who provided devices for dataset collection and feedback for user guidance improvement.

\balance


\bibliographystyle{IEEEtran}
\bibliography{ref}

\end{document}